\documentclass[conference]{IEEEtran}
\IEEEoverridecommandlockouts
\usepackage{cite}
\usepackage{amsmath,amssymb,amsfonts}
\usepackage{algorithmic}
\usepackage{graphicx}
\usepackage{textcomp}
\usepackage{xcolor}
\usepackage[margin=1.5cm]{geometry}
\usepackage{graphicx,subcaption,lipsum}
\def\BibTeX{{\rm B\kern-.05em{\sc i\kern-.025em b}\kern-.08em
    T\kern-.1667em\lower.7ex\hbox{E}\kern-.125emX}}
\begin{document}

\title{Automating grapevine LAI features estimation
with UAV imagery and machine learning\\

\thanks{The work described in the present paper has been developed within the project entitled \textit{VIRMA: Velivolo Intelligente Robotico per il Monitoraggio Agro-Ambientale} - funded by the MUR within the framework of the Fund for the promotion and development of policies of the National Research Program (PNR), in accordance with EU Regulation no. 241/2021 and with the PNRR 2021-2026. The sole responsibility of the issues treated in the present paper lies with the authors.}
}

\author{
\IEEEauthorblockN{Muhammad Waseem Akram}
\IEEEauthorblockA{\begin{minipage}[t]{0.33\textwidth}
\centering
\textit{TeCIP Institute} \\
\textit{Scuola Superiore Sant'Anna}\\
Pisa, Italy \\
muhammadwaseem.akram@santannapisa.it
\end{minipage}}
\and
\IEEEauthorblockN{Marco Vannucci}
\IEEEauthorblockA{\begin{minipage}[t]{0.33\textwidth}
\centering
\textit{TeCIP Institute} \\
\textit{Scuola Superiore Sant'Anna}\\
Pisa, Italy \\
marco.vannucci@santannapisa.it
\end{minipage}}
\and
\IEEEauthorblockN{Giorgio Buttazzo}
\IEEEauthorblockA{\begin{minipage}[t]{0.33\textwidth}
\centering
\textit{TeCIP Institute} \\
\textit{Scuola Superiore Sant'Anna}\\
Pisa, Italy \\
giorgio.buttazzo@santannapisa.it
\end{minipage}}
\and
\IEEEauthorblockN{Valentina Colla}
\IEEEauthorblockA{\begin{minipage}[t]{0.33\textwidth}
\centering
\textit{TeCIP Institute} \\
\textit{Scuola Superiore Sant'Anna}\\
Pisa, Italy \\
valentina.colla@santannapisa.it
\end{minipage}}
\and
\IEEEauthorblockN{Stefano Roccella}
\IEEEauthorblockA{\begin{minipage}[t]{0.33\textwidth}
\centering
\textit{Biorobotics Institute} \\
\textit{Scuola Superiore Sant'Anna}\\
Pisa, Italy \\
stefano.roccella@santannapisa.it
\end{minipage}}
\and
\IEEEauthorblockN{Andrea Vannini}
\IEEEauthorblockA{\begin{minipage}[t]{0.33\textwidth}
\centering
\textit{Biorobotics Institute} \\
\textit{Scuola Superiore Sant'Anna}\\
Pisa, Italy \\
andrea.vannini@santannapisa.it
\end{minipage}}
\and
\IEEEauthorblockN{Giovanni Caruso}
\IEEEauthorblockA{\begin{minipage}[t]{0.33\textwidth}
\centering
\textit{Department of Agriculture Food and Environment} \\
\textit{Università di Pisa}\\
Pisa, Italy \\
giovanni.caruso@unipi.it
\end{minipage}}
\and
\IEEEauthorblockN{Simone Nesi}
\IEEEauthorblockA{\begin{minipage}[t]{0.33\textwidth}
\centering
\textit{Department of Agriculture Food and Environment} \\
\textit{Università di Pisa}\\
Pisa, Italy \\
simone.nesi@phd.unipi.it
\end{minipage}}
\and
\IEEEauthorblockN{Alessandra Francini}
\IEEEauthorblockA{\begin{minipage}[t]{0.33\textwidth}
\centering
\textit{Institute of Crop Science} \\
\textit{Scuola Superiore Sant'Anna}\\
Pisa, Italy \\
alessandra.francini@santannapisa.it
\end{minipage}}
\and
\IEEEauthorblockN{Luca Sebastiani}
\IEEEauthorblockA{\begin{minipage}[t]{0.33\textwidth}
\centering
\textit{Institute of Crop Science} \\
\textit{Scuola Superiore Sant'Anna}\\
Pisa, Italy \\
luca.sebastiani@santannapisa.it
\end{minipage}}
}

\maketitle

\begin{abstract}
The leaf area index determines crop health and growth. Traditional methods for calculating it are time-consuming, destructive, costly, and limited to a scale. In this study, we automate the index estimation method using drone image data of grapevine plants and a machine learning model. Traditional feature extraction and deep learning methods are used to obtain helpful information from the data and enhance the performance of the different machine learning models employed for the leaf area index prediction. The results showed that deep learning based feature extraction is more effective than traditional methods. The new approach is a significant improvement over old methods, offering a faster, non-destructive, and cost-effective leaf area index calculation, which enhances precision agriculture practices.
\end{abstract}

\begin{IEEEkeywords}
Leaf area index, Crop health, Growth estimation, Drone imagery, Precision agriculture, Automated agriculture, Machine learning, Deep learning, Feature extraction
\end{IEEEkeywords}

\section{Introduction}
Global climate challenges and complex field conditions necessitate innovative approaches to enhancing agricultural yields and ensuring food safety. Effective agricultural management heavily relies on rapid, accurate, timely, and non-destructive monitoring of the Leaf Area Index (LAI), a crucial vegetation characteristic closely linked to crop development and influenced by various climatic factors such as atmospheric temperature, precipitation, and solar radiation.
Traditional methods \cite{fang2019overview} for determining LAI, including destructive sampling and manual measurements, are labor-intensive and limited in scale. Conversely, adopting remote sensing technologies \cite{sishodia2020applications} have surged, since they offer promising alternatives. These technologies, particularly satellite imagery and unmanned aerial vehicle (UAV) based remote sensing, enhance yield estimations and monitoring crop growth. For instance, \textit{HJ-1A} satellite remote sensing data \cite{wang2016dynamic} have been used to assess the LAI and biomass of various crops. Estimates of winter wheat yield and other crops like cotton and soybean have been derived from remote sensing images captured by the \textit{ERTS-1} satellite operated by Xie et al. \cite{xie2021integration}. Despite their potential, high-quality satellite images are often expensive and challenged by limitations such as poor geographical and temporal resolution, real-time availability, and atmospheric sensitivity. Moreover, while satellites like Landsat offer extensive coverage, their optical sensors are hindered by a minimum repeat cycle of 16 days, and adverse weather conditions can compromise image accuracy.

Hyperspectral and multispectral imaging technologies via UAVs provide a wide spectral range and high resolution as a more adaptable and cost-effective alternative \cite{sishodia2020applications}. However, those solutions face widespread adoption obstacles due to complexity of data processing and high equipment costs. In developing countries, such challenges are particularly pronounced with small-scale farming. UAV systems equipped with RGB cameras have become a compelling choice due to their superior spatial resolution and more manageable costs, making them particularly suitable for agricultural surveillance in resource-limited settings.
Integrating machine learning (ML) and deep learning (DL) algorithms has significantly advanced the analysis of agricultural remote sensing data \cite{attri2023review,liakos2018machine}. For example,  Li et al. \cite{li2019combining} uses UAV-based photographs to estimate rice LAI via multiple models. Another Wittstruck et al. \cite{wittstruck2022estimating} tested DL models for wheat LAI inversion, and Han et al. \cite{han2019modeling} explore ML techniques to predict maize above-ground biomass. By effectively modeling the non-linearity and heterogeneity in extensive image data, ML and DL algorithms are excellent tools to estimate crop growth. Advanced feature extraction techniques, whether traditional or derived from pre-trained DL models, greatly enhance the efficacy of ML models, providing more comprehensive and informative data representations, as well as facilitating better learning and generalization \cite{attri2023review, Colla2020}.

The study presented in this paper is aimed at comparing various ML models and feature extraction technique combinations to assess their effectiveness in LAI estimation from crop images captured by UAVs. Through experimental validation, we intend to demonstrate the benefits of sophisticated feature extraction methods, particularly those utilizing DL techniques. The objective of this study is to highlight how these technologies can improve the accuracy and efficiency of ML models in the agriculture sector. It offers a quick, cost-effective, and scalable method for automated LAI estimation, enhancing agricultural management and food safety in the face of global climate variability.

\section{Materials and Method}

\subsection{Data collection}

\begin{figure}[ht]
  \centering
  \begin{subfigure}{\columnwidth}
    \centering
    \includegraphics[width=.9\linewidth]{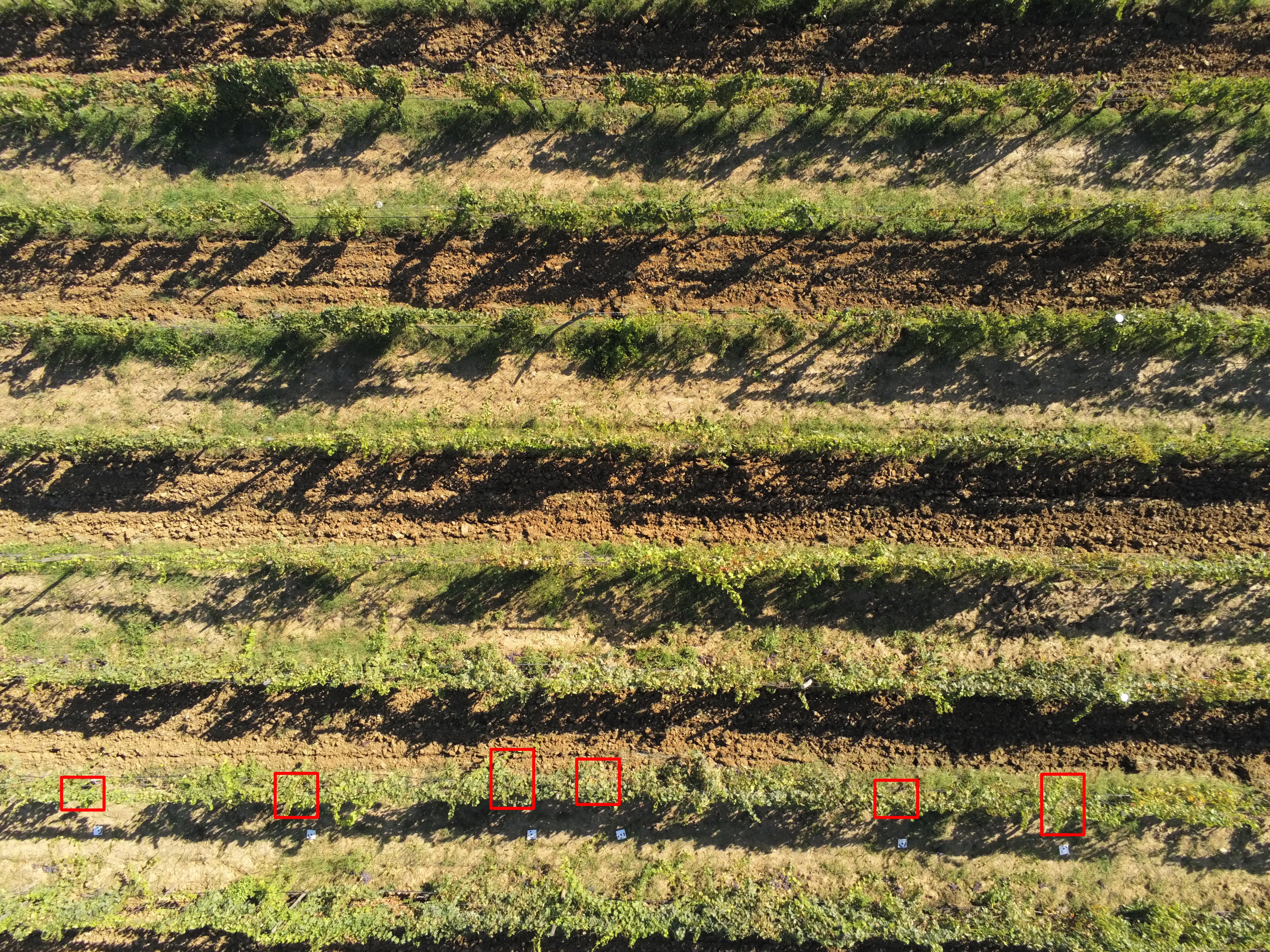}
    \caption*{(a)} % Displays only "a"
  \end{subfigure}
  
  \vspace{1em} % Adjust vertical space between images if needed
  
  \begin{subfigure}{\columnwidth}
    \centering
    \includegraphics[width=.9\linewidth]{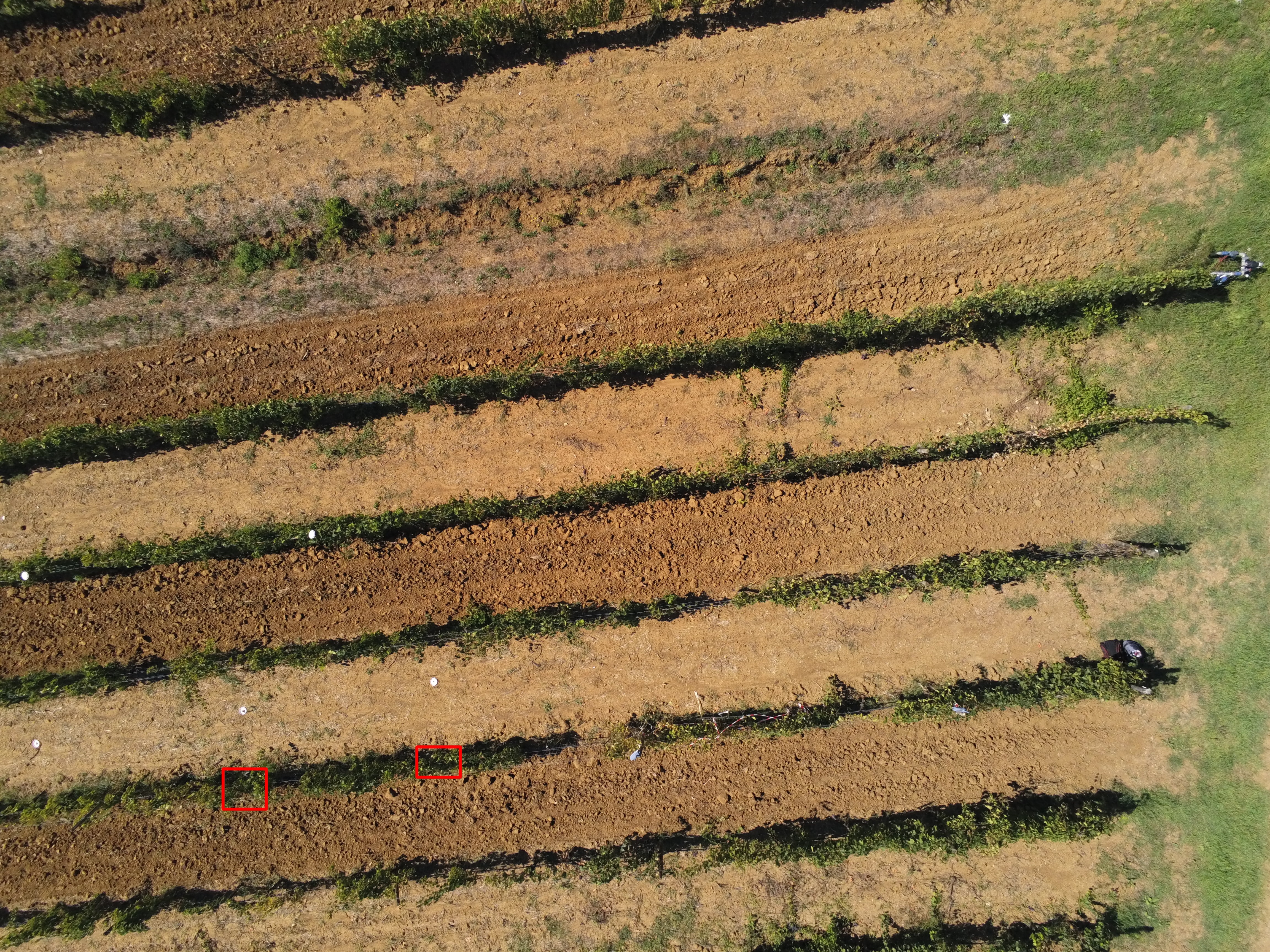}
    \caption*{(b)} % Displays only "b"
  \end{subfigure}
  
  \caption{Examples a and b are annotated images with bounding boxes showing individual plants identified during the annotation process. Each bounding box corresponds to a plant subjected to LAI measurement.}
  \label{fig:bounding_boxes}
\end{figure}

\begin{figure*}[ht]
  \centering
  \begin{subfigure}{.2\textwidth}
    \centering
    \includegraphics[width=.9\linewidth]{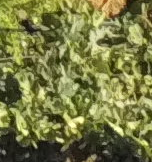}
    \caption*{(a)}
  \end{subfigure}%
  \begin{subfigure}{.2\textwidth}
    \centering
    \includegraphics[width=.9\linewidth]{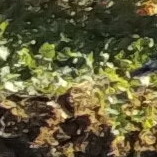}
    \caption*{(b)}
  \end{subfigure}
  \begin{subfigure}{.2\textwidth}
    \centering
    \includegraphics[width=.9\linewidth]{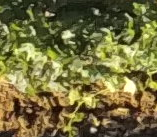}
    \caption*{(c)}
  \end{subfigure}
  \begin{subfigure}{.2\textwidth}
    \centering
    \includegraphics[width=.9\linewidth]{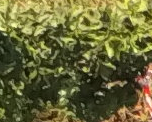}
    \caption*{(d)}
  \end{subfigure}
  \caption{Examples a,b,c and d are images showing individual plants}
  \label{fig:crop}
\end{figure*}

The dataset used for ML training and validation contains images of the investigated individual grapevine plants captured by the drone and their LAI values. The plants were identified by means of a marker placed on the ground that allowed their identification. The LAI was estimated using a non-destructive canopy analysis system (SunScan SS1-R3-BF3, Delta-T Devices, Cambridge, UK) as reported in \cite{caruso2023role}. The images of the individual plants were obtained from the aerial images captured by the drone by means of an annotation process using commercial software that allows generating the bounding box for each plant subjected to  measurement and to associate the corresponding LAI measured value. (See Fig.~\ref{fig:bounding_boxes} for examples of annotated images with bounding boxes highlighting individual plants). This process, currently manual, may be automated in future by using of unique markers.Our dataset consists of 498 images, each associated to a multiple plants and a LAI value. After extracting each plants image the final dataset consist of 1469 images. (See Fig.~\ref{fig:crop} examples of images showing individual plants extracted from the dataset). LAI values are normally distributed with mean value $\eta=0.96$, standard deviation $\sigma=0.67$, minimum value of $\mu=0.24$, maximum value of $M=3.17$.

\subsection{Methodology}

\begin{figure*}[htbp] % Starred version to span two columns
    \centering
    \includegraphics[width=0.95\linewidth]{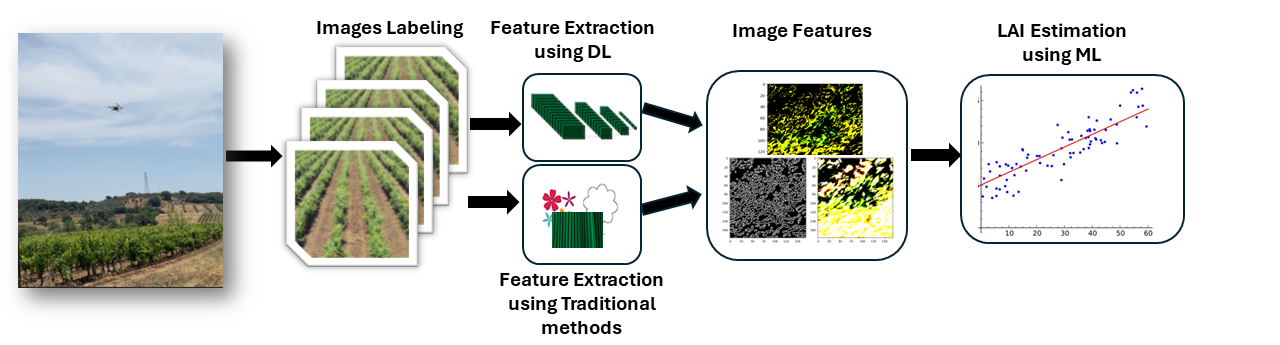}
    \caption{Overview of the Leaf Area Estimation process using drone images and Machine Learning approaches}
    \label{fig:meth}
\end{figure*}

Figure~\ref{fig:meth} depicts the working pipeline of our proposed method for estimating LAI from a dataset \( \mathcal{D} = \{I_1, I_2, \ldots, I_n\} \) consisting of \( n \) images, each associated with a plant within the crop. Features \( \mathcal{F}_k \) are extracted from each image using advanced image processing techniques, denoted by
\begin{equation}
    \mathcal{F}_k = \text{ExtractFeatures}(I_k).
\end{equation}
These features feed into various machine learning (ML) models to accurately predict the LAI, represented by
\begin{equation}
    \text{LAI}_k = \text{EstimateLAI}(\mathcal{F}_k).
\end{equation}
Features are crucial for improving the performance of ML models. We employed three different feature extraction methods:
\begin{enumerate}
    \item Green Area Features Extraction using an Edge Detection Pipeline.
    \item Feature vocabulary development.
    \item Features extraction using a pre-trained deep learning (DL) model.
\end{enumerate}
\subsection{Green Area Features Extraction using an Edge Detection}
Green Area Features Extraction using an Edge Detection \cite{yuan2020extraction} pipeline includes LAI estimation from digital images through a sequence of sophisticated image processing techniques to highlight and isolate vegetation. \begin{itemize}
    \item Firstly, images are thresholded to enhance contrast:
    \begin{itemize}
        \item Pixel values above 50 are set to 255, highlighting key features crucial for vegetation detection.
    \end{itemize}
    \item Images are then converted to the HSV color space:
    \begin{itemize}
        \item This conversion is used to effectively segment green hues via a specific color mask that filters out non-green colors, isolating areas essential for plant analysis.
    \end{itemize}
    \item A Gaussian blur is applied to the green regions:
    \begin{itemize}
        \item This step reduces noise and enhances clarity, facilitating precise edge detection.
    \end{itemize}
    \item Edge detection is performed using Canny’s method:
    \begin{itemize}
        \item Threshold values of 50 and 150 are used for this method.
    \end{itemize}
\end{itemize}
 These steps systematically enhance the raw imagery, emphasizing vegetation edges and contours, which are crucial to accurately measure leaf coverage relative to the ground area, thus providing a reliable LAI value for environmental and agricultural assessment. The reported values for all images are consistent and correspond to the best results. 
\subsection{Feature vocabulary development}
Feature vocabulary development is utilized to combine different image attributes \cite{li2019combining}. LAI estimation from plant imagery benefits from integrating color, shape, and texture features extracted using Color Histograms, Hu Moments, and Haralick Textures. The color histogram for each image,
\begin{equation}
    \text{H}_c(i) = \sum_{p \in I} [c(p) = i],
\end{equation}
assesses the canopy color distribution, particularly focusing on the green spectrum, to estimate leaf coverage density. Hu Moments, resistant to image transformations such as scaling and rotation, quantify geometric features, distinguishing between overlapping leaves and canopy gaps, which is essential for accurate LAI calculation.

Texture features are derived from the Gray-Level Co-occurrence Matrix (GLCM):
\begin{equation}
    \text{GLCM}(i,j) = \sum_{p \in I} [I(p) = i \land I(q) = j],
\end{equation}
From this matrix, measures such as Contrast,
\begin{equation}
    \text{Contrast} = \sum_{i, j} (i-j)^2 \cdot \text{GLCM}(i, j),
\end{equation}
Energy,
\begin{equation}
    \text{Energy} = \sum_{i, j} [\text{GLCM}(i, j)]^2,
\end{equation}
and Homogeneity,
\begin{equation}
    \text{Homogeneity} = \sum_{i, j} \frac{\text{GLCM}(i, j)}{1 + (i-j)^2},
\end{equation}
provide insights into the canopy textural patterns. These metrics, reflecting leaf density and arrangement, enhance the LAI estimation process, allowing for a comprehensive analysis beyond what single indexes offer.

\subsection{Pre-trained model to extract features from images}
Recent advancements in DL that have shown remarkable results in agriculture \cite{attri2023review, liakos2018machine}, we used a pre-trained model to extract features from images. Training a powerful DL model from scratch requires significant amounts of data, time, and resources. Pre-trained models are a more practical option. Features extraction includes a pre-trained ResNet50 model that offers a powerful method to analyse complex image data, such as for LAI estimation \cite{attri2023review}. Leveraging the features learned from extensive visual data by the ResNet model allows capturing detailed characteristics such as leaf texture, shape, and health, which are crucial to accurately determine the LAI. This method significantly enhances the efficiency and robustness of LAI estimation, offering superior performance over traditional image processing techniques by providing a comprehensive understanding of vegetation features critical for ecological and agricultural assessment.

\subsection{Machine Learning Models}

Three different machine learning regression models, including Linear Regression (LR), Support Vector Machines (SVM), and Random Forest (RF), were used to estimate the Leaf Area Index (LAI) from the features extracted through the methods mentioned above. 
\subsubsection{Linear Regression (LR)}
LR serves as one of the most basic and interpretable methods for regression analysis, assuming a direct linear relationship between the predictor variables and the target variable, LAI. The prediction model is defined by the equation:

\begin{equation}
    \hat{y} = \beta_0 + \beta_1 x,
\end{equation}

where \(\hat{y}\) is the predicted value of LAI, \(\beta_0\) is the intercept, and \(\beta_1\) is the coefficient representing the weight of the feature \(x\). Despite its simplicity, LR provides valuable insights into how each feature influences the target, making it particularly useful when the relationships are expected to be linear or nearly linear. However, LR's predictive power can be limited in the presence of non-linear interactions or when the data exhibit complex patterns. Therefore, while LR is efficient and computationally inexpensive, it may not fully capture the intricacies of LAI variability unless the linearity assumption holds.
\subsubsection{Support Vector Machines (SVM)}

SVM extend beyond linear regression by introducing non-linear modeling capabilities through the use of kernel functions. The primary advantage of SVM is its ability to handle high-dimensional data and complex, non-linear relationships. The SVM regression model, or Support Vector Regression (SVR), operates by mapping the input data into a higher-dimensional feature space using a kernel function, thereby allowing the model to fit more complex functions to the data. The SVR model can be expressed as:

\begin{equation}
    f(x) = \mathbf{w} \cdot \phi(x) + b,
\end{equation}

where \(\phi(x)\) denotes the transformation function (kernel), \(\mathbf{w}\) represents the model weights, and \(b\) is the bias term. The Radial Basis Function (RBF) kernel was utilized in this study due to its proficiency in modeling non-linear patterns, which are common in biological and ecological datasets. The RBF kernel effectively measures the similarity between data points based on their distance, making it well-suited for scenarios where relationships are not straightforward. SVM is particularly beneficial when the data is complex and contains noise, as it aims to minimize prediction error while maintaining model robustness.

\subsubsection{Random Forest (RF)}

RF is an ensemble learning technique that combines the outputs of multiple decision trees to improve the prediction accuracy and stability of the model. Unlike individual decision trees, which can be prone to overfitting, RF mitigates this by averaging the results of numerous trees, each constructed from random subsets of the data and features. The RF regression model is expressed mathematically as:

\begin{equation}
    \text{LAI} = \frac{1}{B} \sum_{b=1}^{B} t_b(x),
\end{equation}

where \(B\) is the number of decision trees, and \(t_b(x)\) represents the prediction made by the \(b\)-th tree. For this study, the RF model was configured with 100 trees, optimizing performance while maintaining computational efficiency. This ensemble approach is particularly effective in handling large datasets with high dimensionality and complex interactions among features. RF provides several advantages, such as its ability to model non-linear relationships and automatically handle missing data, making it highly versatile and powerful for predictive modeling.

\subsection{Model Implementation and Parameter Tuning}

These models provide robust, scalable methods for precise LAI calculations \cite{liakos2018machine}. We employed the different parameters specified in the Scikit-Learn library to train the models. This approach involved using the Radial Basis Function (RBF) kernel for the SVM and configuring the RF model to operate with 100 trees, among other settings.

\section{Results and Discussion}
Table \ref{Tab:LAI} compares 3 different ML models, i.e. LR, SVM, and RF applied to 3 feature extraction methods for LAI estimation. Performances are quantified in terms of Mean Squared Error (MSE), Mean Absolute Error (MAE) and Mean Absolute Percentage Error (MAPE). The training dataset includes overlapping leaves and intricate plant structures, thus posing significant hurdles for feature extraction and model training.

The achieved results show that the combination of SVM with ResNet outperforms other models, as it shows the lowest MSE (0.21) and MAE (0.32) values. This proofs a strong fit to \textit{critical} data, showing the superior capabilities of DL algorithms in dealing with the complexity of overlapping leafs and variable plant orientation. 
\begin{table}[htbp]
\caption{Feature Extraction Methods vs. Models for LAI Estimation}\label{Tab:LAI}
  \centering \small
  \begin{tabular}{|c|c|c|c|c|}
    \hline
    \textbf{Feature Extraction Method} & \textbf{Model} & \textbf{MSE} & \textbf{MAE} & \textbf{MAPE} \\
    \hline
    ResNet & LR & 0.55 & 0.59 & 78\% \\
    \hline
    ResNet & \textbf{SVM} & \textbf{0.21} & \textbf{0.32} & \textbf{34\%} \\
    \hline
    ResNet & \textbf{RF} & \textbf{0.23} & \textbf{0.35} & \textbf{34\%} \\
    \hline
    Vocabulary Development & LR & 0.58 & 0.47 & 51\% \\
    \hline
    Vocabulary Development & SVM & 0.39 & 0.45 & 59\% \\
    \hline
    Vocabulary Development & RF & 0.31 & 0.39 & 40\% \\
    \hline
    Green Area & LR & 0.28 & 0.38 & 43\% \\
    \hline
    Green Area & SVM & 0.46 & 0.58 & 40\% \\
    \hline
    Green Area & RF & 0.28 & 0.39 & 43\% \\
    \hline
  \end{tabular}
  \normalsize
\end{table}

The LR model using Vocabulary Development features shows the highest MSE value (0.58), which indicates a scarce predictive accuracy and a tendency to underfitting, most likely due to the method's difficulties in capturing nuanced aspects within complex images.
RF perform consistently across all feature methods, indicating resilience against various feature sets. Notably, when combined to Green Area feature extraction method, RF obtains an MSE comparable to ResNet features, demonstrating that simpler, domain-specific approaches can sometimes outperform more advanced DL techniques.

The variable LR performance across multiple feature sets demonstrates its sensitivity to the adopted feature type, with Green Area features producing the smallest errors. This could indicate that Green Area features are better at capturing linear relationships in data, which are easier for LR to model.
When SVM and RF are paired with ResNet and Green Area feature extraction, both perform well in LAI estimation. SVM great performance with ResNet features is most likely due to its ability to handle complex data structures. Future improvements could include increasing the dataset size and DL fine-tuning to boost model accuracy even further. Meanwhile, our methodology provides agricultural practitioners with practical insights, allowing them to track growth effectively even with small data sets. This makes our approach extremely relevant for both academic research and practical farming applications, ensuring that it remains accessible and valuable at many levels of agricultural management.

\section{Conclusion and Future Work}
This paper presented a comparative evaluation of various ML models and feature extraction techniques to assess their effectiveness in LAI estimation from crop images captured by a drone. The achieved results show that integrating advanced DL techniques and traditional feature extraction methods significantly enhances the accuracy of LAI estimation. While DL models like ResNet excel in processing complex data, simpler methods such as Green Area feature extraction also proved to be effective, offering flexibility for different agricultural settings. Future work will focus on expanding the dataset and fine tuning the DL models to optimize their performance across diverse agricultural landscapes. Moreover, exploring automation of data processing and integrating real-time data analysis could provide more dynamic and accessible solutions for precision agriculture.

\begin{footnotesize}

% IF YOU USE BIBTEX,
% - DELETE THE TEXT BETWEEN THE TWO ABOVE DASHED LINES
% - UNCOMMENT THE NEXT TWO LINES AND REPLACE 'Name_Of_Your_BibFile'

\bibliographystyle{unsrt}
\bibliography{references}

\begin{thebibliography}{10}

\bibitem{fang2019overview}
Hongliang Fang, Frederic Baret, Stephen Plummer, and Gabriela Schaepman-Strub.
\newblock An overview of global leaf area index (lai): Methods, products, validation, and applications.
\newblock {\em Reviews of Geophysics}, 57(3):739--799, 2019.

\bibitem{sishodia2020applications}
Rajendra~P Sishodia, Ram~L Ray, and Sudhir~K Singh.
\newblock Applications of remote sensing in precision agriculture: A review.
\newblock {\em Remote sensing}, 12(19):3136, 2020.

\bibitem{wang2016dynamic}
Jing Wang, Jingfeng Huang, Ping Gao, Chuanwen Wei, and Lamin~R Mansaray.
\newblock Dynamic mapping of rice growth parameters using hj-1 ccd time series data.
\newblock {\em Remote Sensing}, 8(11):931, 2016.

\bibitem{xie2021integration}
Yi~Xie and Jianxi Huang.
\newblock Integration of a crop growth model and deep learning methods to improve satellite-based yield estimation of winter wheat in henan province, china.
\newblock {\em Remote Sensing}, 13(21):4372, 2021.

\bibitem{attri2023review}
Ishana Attri, Lalit~Kumar Awasthi, Teek~Parval Sharma, and Priyanka Rathee.
\newblock A review of deep learning techniques used in agriculture.
\newblock {\em Ecological Informatics}, page 102217, 2023.

\bibitem{liakos2018machine}
Konstantinos~G Liakos, Patrizia Busato, Dimitrios Moshou, Simon Pearson, and Dionysis Bochtis.
\newblock Machine learning in agriculture: A review.
\newblock {\em Sensors}, 18(8):2674, 2018.

\bibitem{li2019combining}
Songyang Li, Fei Yuan, Syed~Tahir Ata-UI-Karim, Hengbiao Zheng, Tao Cheng, Xiaojun Liu, Yongchao Tian, Yan Zhu, Weixing Cao, and Qiang Cao.
\newblock Combining color indices and textures of uav-based digital imagery for rice lai estimation.
\newblock {\em Remote Sensing}, 11(15):1763, 2019.

\bibitem{wittstruck2022estimating}
Lucas Wittstruck, Thomas Jarmer, Dieter Trautz, and Bj{\"o}rn Waske.
\newblock Estimating lai from winter wheat using uav data and cnns.
\newblock {\em IEEE Geoscience and Remote Sensing Letters}, 19:1--5, 2022.

\bibitem{han2019modeling}
Liang Han, Guijun Yang, Huayang Dai, Bo~Xu, Hao Yang, Haikuan Feng, Zhenhai Li, and Xiaodong Yang.
\newblock Modeling maize above-ground biomass based on machine learning approaches using uav remote-sensing data.
\newblock {\em Plant methods}, 15:1--19, 2019.

\bibitem{Colla2020}
Valentina Colla, Costanzo Pietrosanti, Enrico Malfa, and Klaus Peters.
\newblock Environment 4.0: How digitalization and machine learning can improve the environmental footprint of the steel production processes.
\newblock {\em Materiaux et Techniques}, 108(5-6), 2020.

\bibitem{caruso2023role}
Giovanni Caruso, Giacomo Palai, Letizia Tozzini, Claudio D'Onofrio, and Riccardo Gucci.
\newblock The role of lai and leaf chlorophyll on ndvi estimated by uav in grapevine canopies.
\newblock {\em Scientia Horticulturae}, 322:112398, 2023.

\bibitem{yuan2020extraction}
Weitao Yuan, Wangle Zhang, Zhongping Lai, and Jingxiong Zhang.
\newblock Extraction of yardang characteristics using object-based image analysis and canny edge detection methods.
\newblock {\em Remote Sensing}, 12(4):726, 2020.

\end{thebibliography}

\end{footnotesize}

\end{document}